\documentclass[conference]{IEEEtran}
\IEEEoverridecommandlockouts

\usepackage{enumitem}

\makeatletter
\let\NAT@parse\undefined
\makeatother

\usepackage[colorlinks=true]{hyperref}
\usepackage{url}

\usepackage{amsmath}
\usepackage{amssymb}

\usepackage{flushend}

\usepackage[capitalize]{cleveref}

\usepackage[acronym]{glossaries}
\newacronym{llm}{LLM}{Large Language Model}
\newacronym{vlm}{VLM}{Vision Language Model}
\newacronym{lora}{LoRA}{Low-Rank Adaptation}
\newacronym{ap}{AP}{Average Precision}
\newacronym{kp-mse}{KP-MSE}{Keypoint Mean Squared Error}
\newacronym{iou}{IoU}{Intersection Over Union}
\newacronym{vr}{VR}{Virtual Reality}
\newacronym{gmm}{GMM}{Gaussian Mixture Model}
\newacronym{pca}{PCA}{Principal Component Analysis}

\usepackage{booktabs}
\usepackage[table,dvipsnames]{xcolor}

\usepackage{dirtree}

\usepackage{graphicx}

\usepackage{subcaption}

\usepackage{algorithm}
\usepackage{algorithmic}

\usepackage[acronym]{glossaries}
\newacronym{nocs}{NOCS}{Normalized Object Canonical Space}

\usepackage{xspace}
\makeatletter
\DeclareRobustCommand\onedot{\futurelet\@let@token\@onedot}
\def\@onedot{\ifx\@let@token.\else.\null\fi\xspace}

\makeatother

\renewcommand{\aa}{\mathbf{a}}

\providecommand{\oo}{\mathbf{o}}

\providecommand{\tt}{\mathbf{t}}

\def\BibTeX{{\rm B\kern-.05em{\sc i\kern-.025em b}\kern-.08em
    T\kern-.1667em\lower.7ex\hbox{E}\kern-.125emX}}
\begin{document}

\title{Beyond Static Perception: Integrating Temporal Context into VLMs for Cloth Folding
}

\author{
Oriol Barbany$^{1}$
\hspace{4ex} Adrià Colomé$^{1}$
\hspace{4ex} Carme Torras$^{1}$\\
\normalsize{${}^{1}$Institut de Robòtica i Informàtica Industrial, CSIC-UPC} \\ {\tt\small \{obarbany,acolome,torras\}@iri.upc.edu} \\[.5em]
\url{https://barbany.github.io/bifold}
}

\twocolumn[{%
\renewcommand\twocolumn[1][]{#1} %
\maketitle
\thispagestyle{empty}
\begin{center}
    \centering
    \vspace{-2.5em}
    \captionsetup{type=figure}
    \includegraphics[width=\textwidth]{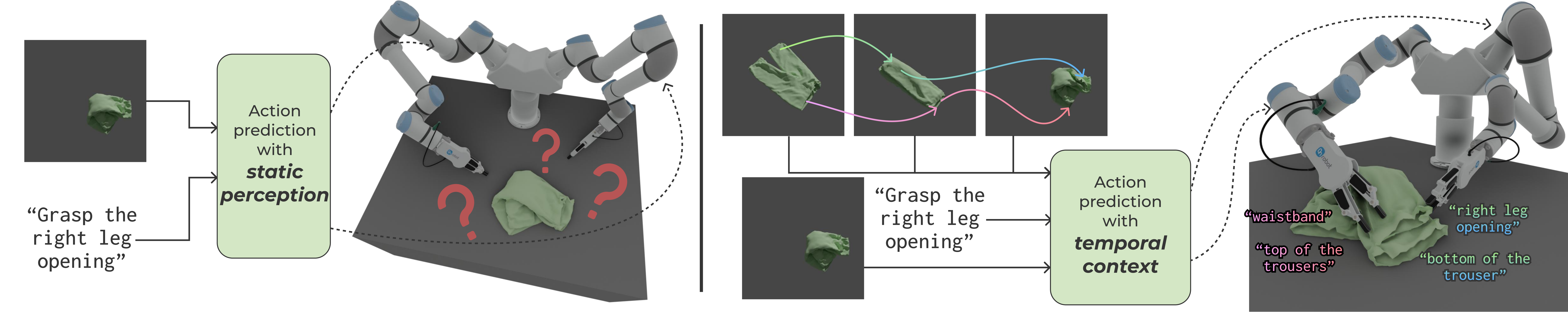}
    \captionof{figure}{\textbf{Motivation}:  In multi-step tasks like cloth folding, the state can become visually ambiguous due to self-occlusions or compounding errors from past actions. Relying solely on the current observation can make the next action ill-defined (left). By incorporating temporal context (right), the model gains critical information enabling more accurate perception and decision-making.
    \looseness=-1
    }
    \label{fig:teaser}
\end{center}
}]

\begin{abstract}
    Manipulating clothes is challenging due to their complex dynamics, high deformability, and frequent self-occlusions. Garments exhibit a nearly infinite number of configurations, making explicit state representations difficult to define. In this paper, we analyze \textit{BiFold} \cite{barbany2025bifold}, a model that predicts language-conditioned pick-and-place actions from visual observations, while implicitly encoding garment state through end-to-end learning. To address scenarios such as crumpled garments or recovery from failed manipulations, BiFold leverages temporal context to improve state estimation. We examine the internal representations of the model and present evidence that its fine-tuning and temporal context enable effective alignment between text and image regions, as well as temporal consistency.
\end{abstract}


\section{Introduction} 
Cloth folding is a fundamental and repetitive task in daily life, present in both domestic and industrial settings. It is a labor-intensive activity that consumes significant time and effort and would greatly benefit from automation. Robotic cloth folding has the potential to free humans from these tedious chores and increase the productivity in sectors such as laundry services, textile manufacturing, and even in-home robotics.
However, achieving robust and versatile robotic cloth folding is challenging due to the inherent complexity of cloth manipulation, including its high deformability, self-occlusions, and diverse material properties, topologies, and textures.

While some robotized tasks can be performed in open-loop or with simple sensing systems, perception plays a key role in deformable object manipulation. As in other vision-based tasks, changes in color distribution due to varying lighting conditions or textures can cause system failures. A standard solution in prior work on cloth manipulation is to discard color information and rely solely on depth images \cite{wang2023trtm,language_deformable_manipulation,mo2022foldsformer,weng2021fabricflownet}. However, depth images obtained from real sensors are noisy and color provides useful state recognition cues.

Most previous works decouple perception from the manipulation policy by explicitly extracting a representation and predicting actions based on it. In most cases, this representation is mesh-based \cite{garment_tracking,chi2021garmentnets,caldarelli23,huang22occlusion_reasoning,lin2022learning}, but alternative approaches include explicit surfaces \cite{barbany24deformable}, Gaussians \cite{longhini2024clothsplatting,zhang2024dynamics}, implicit surfaces \cite{li2024folding,guillard2022udf}, and keypoints \cite{lips24keypoints}, among others. Nonetheless, each has limitations, whether in flexibility to new topologies, resolution for capturing wrinkles or crumpled states, or richness of encoded information for effective manipulation.
\looseness=-1

In this work, we build on the BiFold model \cite{barbany2025bifold}, which adapts a pretrained \gls*{vlm} for language-conditioned bimanual cloth folding. Rather than introducing a new method, this workshop paper presents a deeper analysis of the internal mechanisms learned by BiFold. We explore how the model encodes visual and linguistic information across time, focusing on feature representations, attention patterns, and the role of temporal context. Our goal is to understand how BiFold interprets complex cloth states and leverages history for accurate perception, offering insights that can inform future work in deformable object manipulation.

\section{Methodology}

\begin{figure*}[ht]
    \centering
    \includegraphics[width=\linewidth]{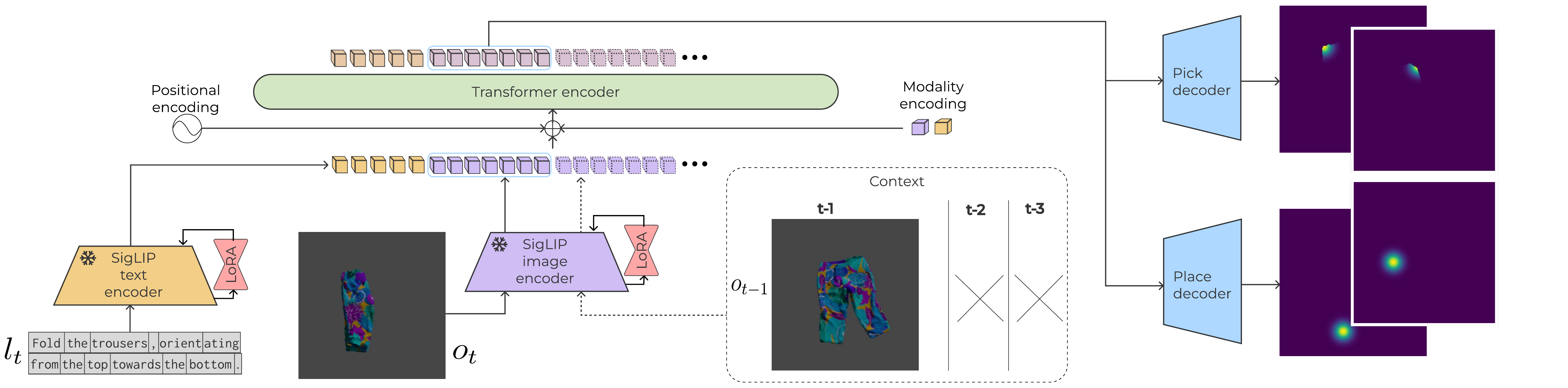}
    \caption{\textbf{BiFold architecture: }BiFold fine-tunes
    a pre-trained
    SigLIP \cite{siglip} with LoRA \cite{hu2022lora} to obtain image and text features. The same image encoder is used for the current and past observations. Each token sequence is prepended with a different learned token, and we add information about the modality and the position inside the sequence by adding a learned embedding and sinusoidal positional encodings. All tokens are concatenated and processed using a transformer encoder \cite{transformers} and the output tokens of the current observation are decoded into probability distributions on the pixel space
    using convolutional networks.}
    \label{fig:architecture}
    \vspace{-1.5em}
\end{figure*}

We focus on bimanual cloth folding starting from flattened configurations with sub-goals specified through natural language instructions. Specifically, we analyze the BiFold model \cite{barbany2025bifold}, which we introduce below. For further details on the model architecture, dataset creation, and comprehensive experimental results, we refer the reader to the original paper.

At each time step $t$, given an RGB image observation $\oo_t$ and a language instruction $\ell_t$, BiFold predicts an action $\aa_t$ using a learned policy $\pi_\theta(\aa_t | \ell_t, \oo_t, \oo_{t-1}, \oo_{t-2}, \dots, \oo_{t-H})$ parametrized by $\theta$. In the base configuration, BiFold sets $H=3$, which we use in our experiments. Crucially, the past observations correspond to \textit{keyframes}, not raw camera frames as in works like RT-1 \cite{rt1}. Since consecutive frames captured at high frequency are often highly redundant, BiFold instead uses observations sampled before the execution of each action, which provide more meaningful temporal context.

To effectively learn from vision and language inputs in a low-data regime, BiFold adapts SigLIP \cite{siglip}, a contrastive \gls*{vlm} similar to CLIP \cite{clip}
composed of independent transformer-based encoders for images and text that produce aligned features.
SigLIP offers high-level semantic representations that are well-suited for language-conditioned action prediction \cite{kim24openvla}, and has demonstrated a low sim-to-real gap in robotic applications \cite{SPOC_2024_CVPR}. SigLIP is adapted using parameter-efficient fine-tuning to learn task-specific features while preserving the generalization capabilities of the foundational model. Specifically, BiFold employs \gls*{lora} \cite{hu2022lora}, a widely adopted method for adapting transformer-based models to downstream tasks \cite{zanella2024low, barbany24multimodal_search, feng2025reflective, kim24openvla}.
This design choice reduces the number of trainable parameters to just 0.29\% of the more than 200 million parameters in the vision and text encoders.

Given the token sequences for the current image, language instruction, and context images, BiFold prepends each sequence with a distinct learned embedding. To encode positional information
, BiFold applies 1D and 2D sinusoidal encodings for text and image tokens, respectively. Additionally, it includes a learned modality embedding to further distinguish between text and image inputs. BiFold then fuses all tokens by concatenating the sequences and applying cross-modal attention layers, following a strategy similar to \cite{octo_2023,rt2}.

The output tokens corresponding to the current observation are then fed into convolutional decoders, which produce probability distributions over the pixel space. BiFold finally samples the pick and place positions for both the left and right arms from the predicted distribution to form the action $\aa_t$, which can be passed to robot-specific controllers to execute actions.

The policy parameters $\theta$ are learned using the BiFold dataset \cite{barbany2025bifold}, which contains language-aligned bimanual folding actions across various garment types. The dataset is constructed through a novel pipeline that extracts discrete actions from the continuous human demonstrations in the VR-Folding dataset \cite{garment_tracking}. Each garment is then canonically aligned within its category, enabling the generation of language instructions based on positions in this normalized space. Notably, the entire procedure is fully automatic, allowing the dataset to scale efficiently to new demonstrations, garment assets, and even unseen categories. More details in the full paper \cite{barbany2025bifold}.

BiFold minimizes the binary cross-entropy loss between the predicted probability distribution and a \gls*{gmm}, where each Gaussian is centered at one of the ground-truth pick or place positions and assigned equal weight. For the pick positions, we restrict the Gaussian's support to the segmentation mask to avoid grasps at positions where there's no cloth. The resulting target mixture is normalized to ensure it defines a valid probability distribution over the pixel space.

\section{Results}

We conduct experiments and provide visualizations that offer insight into the internal behavior of BiFold to support the following claims:

\begin{enumerate}[label=(\roman*)]
    \item Parameter-efficient fine-tuning of \glspl*{vlm} enables extracting more task-specific representations while retaining their robustness and open-vocabulary understanding.
    \item Incorporating a \textit{keyframe} history improves the model’s ability to recognize and interpret complex object states.
\end{enumerate}

\subsection{Quantitative comparison}

\begin{table}[t]
    \caption{\textbf{Quantitative metrics on BiFold dataset 
    \cite{barbany2025bifold}}}
    \resizebox{%
      \ifdim\width>\linewidth
        \linewidth
      \else
        \width
      \fi
    }{!}{
    \begin{tabular}{lcccc}
        \toprule
        & $\text{AP}_{5}$ ($\uparrow$)& $\text{AP}_{10}$ ($\uparrow$)& $\text{AP}_{50}$  ($\uparrow$) & Quantile (\%) ($\uparrow$) \\
        \midrule
        BiFold w/o context & {40.3} & {55.7} & {88.9} & {97.1} \\
        BiFold consecutive & 34.6 & 49.8 & 83.9 & 97.3 \\
        BiFold & \textbf{75.1} & \textbf{92.8} &  \textbf{98.9} & \textbf{98.8} \\
        \bottomrule
    \end{tabular}
    }
    \label{tab:quantitative}
\end{table}

We evaluate the performance of BiFold and two variants of it: without context and with context using the consecutive frames rather than the previous \textit{keyframes} to evaluate (ii).

In \cref{tab:quantitative}, we report the \gls*{ap} at different pixel thresholds and the output quantile, which indicates how likely are the ground truth positions in the predicted pick and place distributions. We can see that BiFold performs best, and its version with consecutive frames achieves, in most cases, worse metrics than using no context at all. We hypothesize this is due to considering uninformative tokens, which can make the model more susceptible to noise and harder to train effectively. Including all past frames introduces redundant or irrelevant visual information, especially in high-frame-rate settings, leading to feature dilution and potentially misaligned attention. Moreover, the increased sequence length demands more computational resources and may hinder learning, as the model needs to distinguish useful signals from a larger pool of mixed-quality tokens. 

\subsection{Implicit feature matching}

\begin{figure}[t]
    \centering
    \captionsetup[subfigure]{labelformat=empty}
    \begin{subfigure}{\linewidth}
        \centering
        \includegraphics[trim={390 650 
        390 140},clip,width=.28\linewidth]{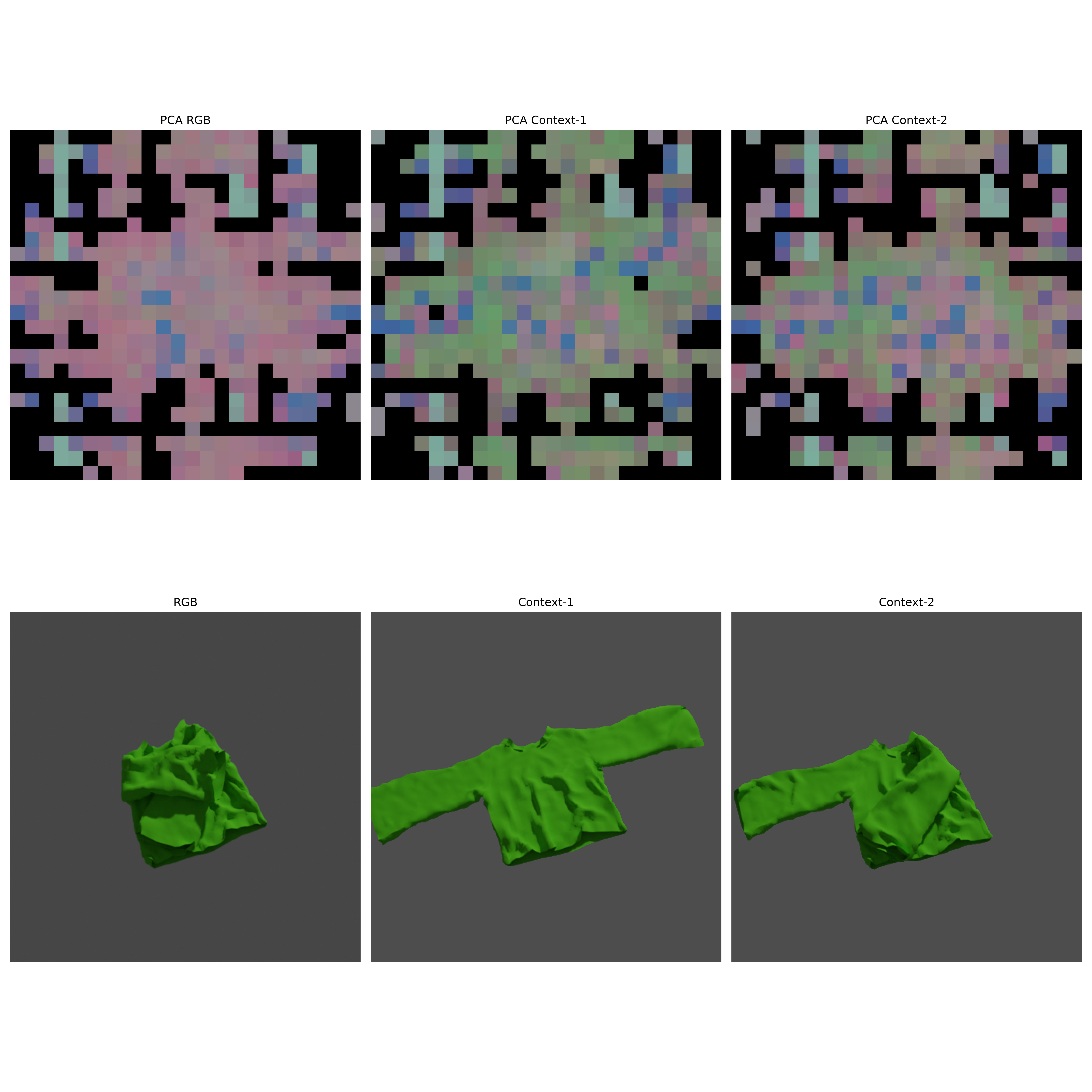}
        \includegraphics[trim={770 650 10 140},clip,width=.28\linewidth]{imgs/pretrained_features.png}
        \includegraphics[trim={10 650 770 140},clip,width=.28\linewidth]{imgs/pretrained_features.png}
        \caption{Pre-trained features}
    \end{subfigure}
    \begin{subfigure}{\linewidth}
        \centering
        \includegraphics[trim={390 650 
        390 140},clip,width=.28\linewidth]{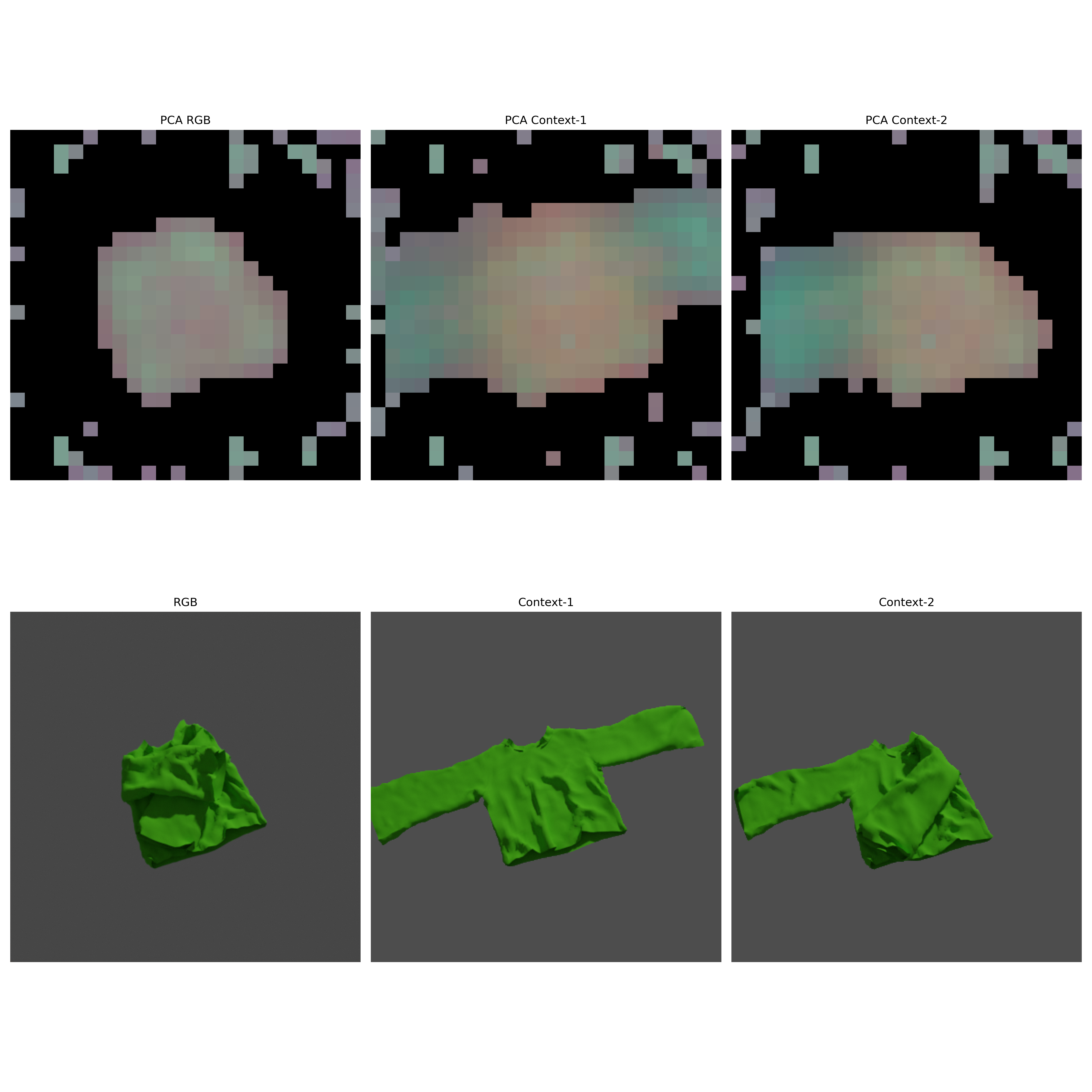}
        \includegraphics[trim={770 650 10 140},clip,width=.28\linewidth]{imgs/finetuned_features.png}
        \includegraphics[trim={10 650 770 140},clip,width=.28\linewidth]{imgs/finetuned_features.png}
        \caption{BiFold features}
    \end{subfigure}
    \begin{subfigure}{\linewidth}
        \centering
        \includegraphics[trim={390 140 390 650},clip,width=.28\linewidth]{imgs/pretrained_features.png}
        \includegraphics[trim={770 140 10 650},clip,width=.28\linewidth]{imgs/pretrained_features.png}
        \includegraphics[trim={10 140 770 650},clip,width=.28\linewidth]{imgs/pretrained_features.png}
        \caption{Observations ($\oo_{t-2}$,$\oo_{t-1}$, and $\oo_{t}$)}
    \end{subfigure}
    \caption{\textbf{Image encoder features: }While pre-trained SigLIP features offer aligned language and visual embeddings, they struggle to capture semantic correspondences across time steps. In contrast, BiFold's fine-tuned version focuses more effectively on the manipulated cloth, enhancing the model's ability to distinguish task-relevant regions critical for folding.}
    \label{fig:pretrained_features}
    \vspace{-1.5em}
\end{figure}

To qualitatively assess (i), we analyze the patch-level features extracted by the image and text encoders. Specifically, we apply \gls*{pca} and visualize the first three components as RGB color maps. Following DINOv2 \cite{oquab2024dinov}, we retain only those patches whose first principal component is non-negative To address (ii) and evaluate the impact of temporal context, we perform \gls*{pca} on the aggregated features from both the current observation and its history of \textit{keyframes}.

As shown in \cref{fig:pretrained_features}, the pre-trained features retained after \gls*{pca} thresholding are not well localized on the garment, and there is a lack of alignment across frames. In contrast, after fine-tuning, the features more effectively isolate the main object from the background, and exhibit semantically meaningful correspondences across frames—such as consistent activation on sleeves, the center of the shirt, and their contours.

\subsection{Attention maps}

\begin{figure}[t]
  \centering
  \begin{subfigure}[b]{0.49\linewidth}
    \centering
    \includegraphics[width=0.48\textwidth]{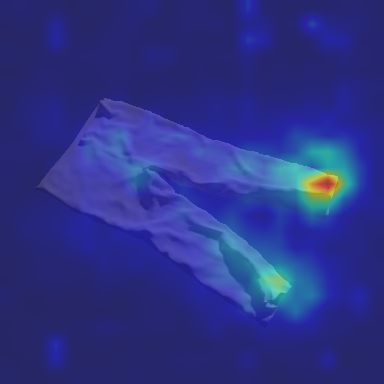}
    \includegraphics[width=0.48\textwidth]{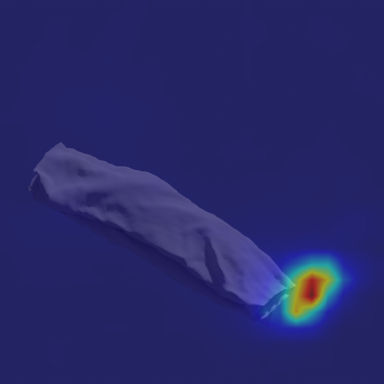}
    \caption{\texttt{"bottom"} in $\oo_{t-1}$ and $\oo_t$}
    \label{fig:att_bottom}
  \end{subfigure}
  \begin{subfigure}[b]{0.49\linewidth}
    \centering
    \includegraphics[width=0.48\textwidth]{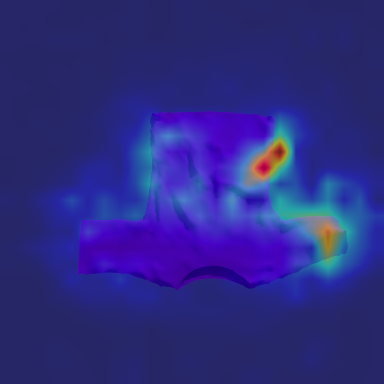}
    \includegraphics[width=0.48\textwidth]{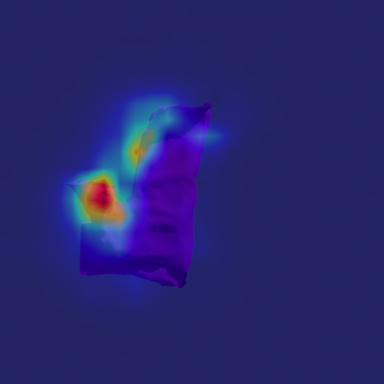}
    \caption{\texttt{"fold"} in $\oo_{t-1}$ and $\oo_t$}
    \label{fig:att_fold}
  \end{subfigure}
  \begin{subfigure}[b]{0.74\linewidth}
    \centering
    \includegraphics[width=0.32\textwidth]{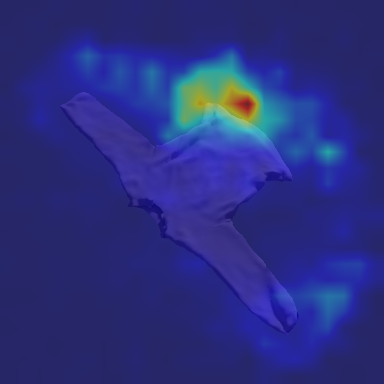}
    \includegraphics[width=0.32\textwidth]{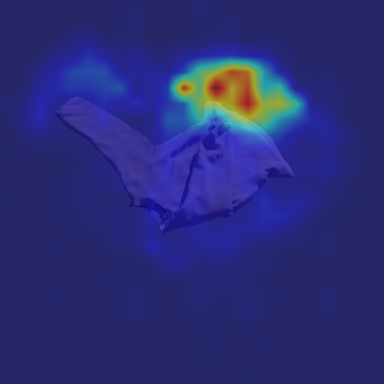}
    \includegraphics[width=0.32\textwidth]{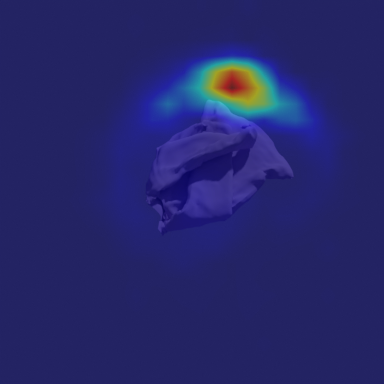}
    \caption{\texttt{"bottom"} in $\oo_{t-2}$, $\oo_{t-1}$ and $\oo_t$}
    \label{fig:att_bottom_3}
  \end{subfigure}
  \hfill
  \begin{subfigure}[b]{0.24\linewidth}
    \centering
    \includegraphics[width=\textwidth]{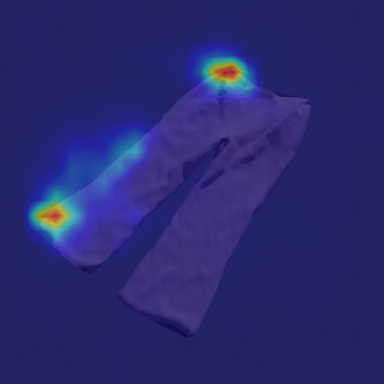}
    \caption{\texttt{"fold"} $\oo_t$}
    \label{fig:att_fold_4}
  \end{subfigure}

  \caption{\textbf{Text$\to$Image attention: }Attention scores from text to image patches to the current observation and \textit{keyframes}.}
  \label{fig:attention}
  \vspace{-1em}
\end{figure}

Before the transformer encoder in \cref{fig:architecture}, the features for the input text and each of the observations are extracted independently. We now analyze how text tokens attend to visual tokens within the transformer encoder, providing insight into how language guides perception. In \cref{fig:attention}, we visualize attention maps over the current observation and all previous \textit{keyframes} (when available) for task-relevant text tokens particularly informative for cloth folding. The token \texttt{"bottom"} attends to the lower regions of garments, while \texttt{"fold"} exhibits bimodal attention corresponding to plausible bimanual pick locations. In both cases, the attention maps reveal consistent spatial focus across temporal frames, supporting our claim (ii) that historical context enhances state recognition.

\section{Discussion}
\label{sec:discussion}

Given the high degrees of freedom of garments and the challenges in explicitly modeling them, BiFold \cite{barbany2025bifold}
implicitly encodes the state using a fine-tuned \gls*{vlm}. We showed that BiFold learns attention patterns aligned with the folding task and focuses on relevant features, effectively segmenting the manipulated cloth without supervision as in \cite{oquab2024dinov}. Although BiFold learns cross-frame mappings without explicit supervision, additional learning signals such as NOCS \cite{nocs} images could further encourage temporal consistency and spatial grounding.

While selecting \textit{keyframes} is straightforward in our setting, this becomes more challenging in manipulation tasks where actions are not easily segmentable or involve many interdependent steps, making intermediate frames potentially valuable. One solution could be to automatically select the most informative frames from a history buffer or allocate tokens proportionally to their importance \cite{yan2025elastictok}.


\clearpage

\bibliographystyle{IEEEtran}
\bibliography{IEEEabrv,bibliography}

\end{document}